\documentclass[conference,a4paper]{APSIPA2018}
\usepackage{multirow}
\usepackage[dvips]{graphicx}
\usepackage{amsmath}
\usepackage[psamsfonts]{amssymb}
\usepackage{amsxtra}
\usepackage{threeparttable}
\usepackage[whole]{bxcjkjatype}

\usepackage{subfigure}
\usepackage{balance}

\begin{document}

\title{Journal Name Extraction from Japanese Scientific News Articles}

\author{
    \IEEEauthorblockN{Masato Kikuchi \quad Mitsuo Yoshida \quad Kyoji Umemura}
    \IEEEauthorblockA{Toyohashi University of Technology, Aichi, Japan\\
    \{m143313@edu, yoshida@cs\}.tut.ac.jp, umemura@tut.jp}
}

\maketitle
\thispagestyle{empty}

\begin{abstract}
In Japanese scientific news articles, although the research results are described clearly, the article's sources tend to be uncited.
This makes it difficult for readers to know the details of the research.
In this paper, we address the task of extracting journal names from Japanese scientific news articles.
We hypothesize that a journal name is likely to occur in a specific context.
To support the hypothesis, we construct a character-based method and extract journal names using this method.
This method only uses the left and right context features of journal names.
The results of the journal name extractions suggest that the distribution hypothesis plays an important role in identifying the journal names.
\end{abstract}

\section{Introduction}

In Japanese scientific news articles, although the research results are described clearly, the article's sources tend to be uncited.
This makes it difficult for readers to know the details of the research.
However, the journal names are usually described in the article.
If we could extract the journal names automatically by using a specific algorithm, we could use the information to find the uncited sources.
In this paper, we extract journal names from Japanese scientific news articles found on news websites and list the results.

Named Entity Recognition (NER) is used to automatically extract named entities, such as journal names, from large documents~\cite{Nadeau:07}.
In NER, supervised learning methods are often used for datasets tagged with named entities.
The named entity dictionary is also effective as a source of external knowledge to identify named entities.
The advantage of the dictionary is its high extraction accuracy, and the possibility of extracting many named entities by preparing a large-scale dictionary that can sufficiently cover all the named entities.
However, such a large-scale dictionary is hard to create manually.
Therefore, bootstrap methods have been proposed in references~\cite{Riloff:99,Collins:99,Yangarber:02} and~\cite{Pantel:06}.
These methods are based on a common approach.
This approach uses a few named entities called seeds as initial training data, and it alternately repeats the extraction of named entities and expands of training data.
When writing the names of foreign journals in Japanese articles, the names are often not translated as individual words, but by applying Japanese characters to the journal name's pronunciations.
This causes variations in spelling by the authors.
Moreover, journal names are diverse.
For the above reasons, it is difficult to manually create a list of the journal names in advance.
Therefore, we extract the journal names based on the bootstrap approach.

Journal names tend to occur in specific contexts since their roles in the articles are decided in advance.
We try to assume that the distribution hypothesis~\cite{Harris:54} is effective in extracting journal names; the hypothesis states that words used in the same context tend to have similar meanings.
We regard all journal names as having similar meanings.
In this paper, we verify the effectiveness of the distribution hypothesis in extracting journal names. 
We do this by constructing a character-based method with only left and right context features as the occurrence pattern of the journal names,
and extracting the journal names using this method.
Many complicated and advanced NER methods exist.
However, to accurately analyze the effect of the distribution hypothesis on the journal name extractions, we construct the simple and generic method, which only uses the left and right context features for journal name extractions.
We extract journal names from Japanese scientific news articles collected from websites.

We hypothesize that a journal name is likely to occur in a specific context.
Our contribution is to support the hypothesis for journal name extractions, 
by constructing the method described in the above and extracting journal names.
The extraction results suggest that the distribution hypothesis plays an important role in identifying journal names.
Furthermore, we also analyze the results of the failed example of the extractions.
The analysis results show that other features are needed in addition to the left and right context features in order to continuously and accurately extract journal names when using the bootstrap approach.
Our future aim is creating a powerful method specialized in journal name extractions.
In this method, we believe that the left and right context features will be effective to extract many journal names.

\section{Related work}

Many bootstrap methods for extracting named entities have been proposed by various studies.
In reference~\cite{Riloff:99}, a two-tiered bootstrapping process is proposed.
The first level of this bootstrapping process extracts unique patterns and domain-specific terms simultaneously.
The second level of this bootstrapping process deletes ambiguous terms from the dictionary.
In reference~\cite{Collins:99}, a method to classify the named entities using their spellings and the context in which they occur is proposed.
In reference~\cite{Yangarber:02}, using the fixed-length contexts around the named entities as contextual patterns is proposed.
Note that the patterns used in this method are only on one side of the left or right contexts.
In reference~\cite{Pantel:06}, a method that extracts the semantic relationships from the general contexts using large-scale resources, such as web resources, is proposed.
This study also adopts a bootstrap method.
In our method, contextual patterns are identified from not only one side, but both sides of the context.

In Japanese NER, context features of both sides are often used as contextual patterns.
In general, the left and right context features of the named entities are extracted with fixed lengths, and the part-of-speech features to which the contexts belong are used.
The input string is divided into analysis units called tokens by a morphological analysis, and the named entity parts are chunked using the part-of-speech features.

If the tokens are used for chunking directly, the named entities that are smaller than the tokens cannot be extracted.
Therefore, reference~\cite{Asahara:03} proposes chunking in character units.
However, when the number of components of a named entity is large, this approach often causes analysis errors because insufficient information is given by chunker.
Reference~\cite{Nakano:04} uses Japanese-based phrases called {\it bunsetsu} as a new feature to solve the above-mentioned problem.
The method in~\cite{Nakano:04} cannot extract named entities containing attached words\footnote{Attached words are the words which cannot be \textit{bunsetsu} by themselves.} and conjugational words\footnote{Conjugational words indicate Japanese verbs, adjectives, and so on.}.

Journal names often have many components and contain attached words and conjugational words.
The method we construct in this paper uses the context of both sides of the named entity but extracts the journal names using a statistical framework instead of a morphological analysis.
In our method, it is possible to extract the journal names that are difficult to extract using the methods mentioned above.

In reference~\cite{Pantel:09}, an approach that uses distribution similarity with all the words as features to recognize named entities is proposed.
However, the feature space of the words is known to be multidimensional and sparse.
We only use bigrams located in both sides of journal names because we believe that the names occur in specific contexts.

\section{A character-based method}

In this section, we describe a character-based method for the journal name extractions.
This method receives news articles as input and extracts the journal names via the bootstrap approach from the body of each article.
Although this method is simple and generic, it eases to analyze the effect of the distribution hypothesis on the journal name extractions.
This method consists of the following three steps:

\noindent
\textbf{Step1. Generating training data: }
first, we divide the article body into string bigram units.
Then, we count and retain the occurrence frequency of each bigram type throughout the dataset.

\begin{figure}[tp]
  \centering
  \includegraphics[width=0.99\linewidth]{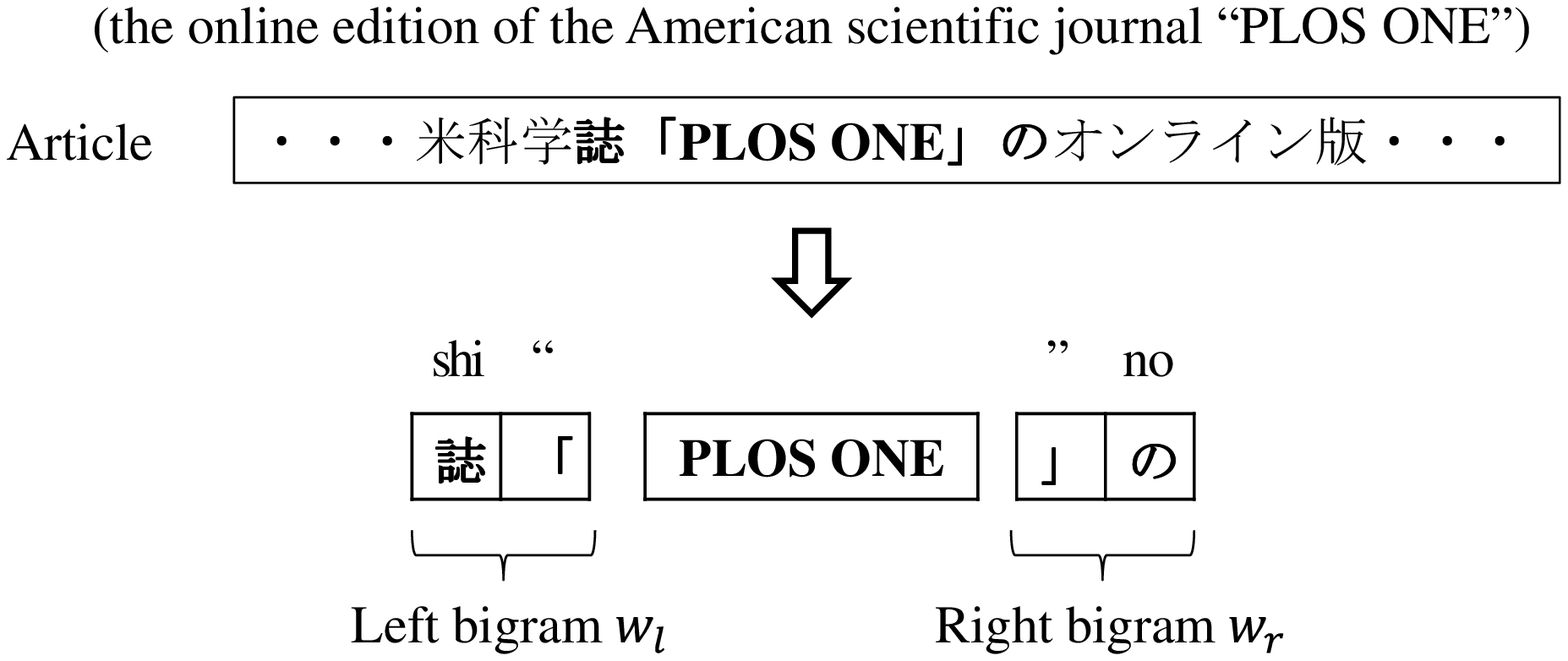}
  \caption{
An example of the journal name extraction and the left / right bigrams.
The journal name is ``PLOS ONE''.
In general, two consecutive Japanese characters are regarded as one bigram.
}
  \label{fig:Exract_bigrams}
\end{figure}

Next, we find all the journal names in the dictionary from the dataset using a longest-match search and extract the names along with their left and right bigrams (Figure~\ref{fig:Exract_bigrams}).
If a journal name is at the beginning or end of an article, then the left or right bigram, respectively, does not exist.
In this case, we give the code {\textless NONE\textgreater \textless NONE\textgreater} to the positions where the bigrams do not exist.
If only one character does not exist, we use the code {\textless NONE\textgreater}.
For each of the left and right bigrams, we count the occurrence frequency for each bigram type.
These frequencies are retained separately for the left and right bigrams.
Note that {\textless NONE\textgreater \textless NONE\textgreater} is excluded from the occurrence frequency counts.

Lastly, we correct the counted frequencies using the Simple Good-Turing method~\cite{Gale:95} for each position of the entire dataset and the journal names' left or right bigrams.
This frequency correction is able to solve the zero frequency problem by improving the performance of the journal name extractions.
This method corrects the frequencies, based on the expected values of the frequencies of the bigrams.
Two estimation approaches, the Turing estimator and the Good-Turing estimator, are used in this method.
In the Turing estimator, the corrected frequency, $GT(r)$, of the bigram that occurs $r$ times is estimated as follows:
\begin{align}
\label{eq:Smoothed frequency}
GT(r) = (r+1) \frac{N_{r+1}}{N_r},
\end{align}
where $N_r$ is the number of bigram types occurring $r$ times in the training data.
$GT(0)$ is estimated by the following formula:
\begin{align}
GT(0) = (0+1) \frac{N_{0+1}}{N_0}.
\end{align}
However, the number of unseen bigram types, $N_0$, cannot be observed from the training data.
Therefore, $N_0$ is estimated by subtracting the number of observed bigram types from the square of the number of observed unigram types~\cite{Manning:99}.
Through the above processes, we can generate training data by the frequency of bigrams seen in each position of the entire dataset and the journal names' left or right bigrams.

\noindent
\textbf{Step2. Journal name extraction: }
first, we specify the range of the length of the journal names to be extracted.
Then, we extract the journal name candidates that have the specified name length and left and right bigrams from the dataset.
In this study, the candidates together with their left and right bigrams are called {\it extraction strings}.
When extracting the strings, we fix the head positions of the candidates at the beginning of the articles.
We then extract the candidates and their left and right bigrams while increasing the length of the candidates by one character within the specified length range.
Once the length of the candidates reaches the maximum range of length or the candidates reach the end of the article, we shift the head positions by one character to the right and start the extraction again.
When the candidates' head positions are at the beginning of the article or the candidates reach the end of the article, there are no left or right bigrams.
In this case, we give the code {\textless NONE\textgreater \textless NONE\textgreater} to the positions where the bigrams do not exist.
If only one character does not exist, we use {\textless NONE\textgreater}.

We calculate the score for the candidates, which are written as $w$. 
The score can be calculated using the geometric mean of the likelihood ratios as shown in Eq. (\ref{eq:Likelihood Ratio}).
\begin{align}
\label{eq:Likelihood Ratio}
S(w)&=(LR_l \times LR_r)^\frac{1}{2} \nonumber \\
&=\left\{\frac{P(w_l \mid Occ_l)}{P(w_l \mid Occ_{all})}  \times \frac{P(w_r \mid Occ_r)}{P(w_r \mid Occ_{all})}\right\}^\frac{1}{2},
\end{align}
where $w_l$ and $w_r$ are the left and right bigrams of $w$.
$Occ_{all}$, $Occ_l$ and $Occ_r$ indicate that the bigrams occur in each position of the entire dataset and left or right of $w$, respectively.
Each likelihood ratio is the ratio of two probabilities.
The denominator is the probability that bigram $w_{*}$ will occur in somewhere in the dataset.
The numerator is the probability that bigram $w_{*}$ will occur in left or right of $w$.
We use the maximum likelihood estimator (MLE) for probability estimation.
We compute the MLE by using observed frequencies.
When the likelihood ratio is larger than one, the bigram tends to occur in left or right of $w$.
However, when the ratio is smaller than one, it tends to occur in somewhere in the dataset.
As the score increases, the candidates are more likely to be a journal name.
The score is the geometric mean of the likelihood ratios.
Therefore, if the probabilities are zero, the score would be zero or infinity.
Specifically, when bigram $w_*$ is not observed in the training data, we cannot calculate the score of the candidate.
To solve this problem, as described in step 1, we assign a small positive value to the zero frequency by using the Simple Good-Turing method.

We extract the top $N$ items in descending order of score and make manual judgments on the correctness.
Then, we add the names we deem correct to the dictionary.

\noindent
\textbf{Step3. Repeat: }
we alternately repeat steps 1 and 2 until a sufficient number of journal names are obtained.
When extracting candidates repeatedly, we do not extract candidates containing only one of the corresponding parentheses: 「」, 『』, （）, and ().

\section{Experiment}

We extracted journal names from Japanese scientific news articles collected from websites and quantitatively evaluated the performance of the method described in section III.
As shown in TABLE~\ref{tab:Seeds}, we used 10 journal names as seeds.
The extracted journal names were between 2 and 50 characters long.
We extracted the top 2,000 journal name candidates, excluding the seeds, in descending order of score.
The performance measures used were precision, recall, and F-measure.
We conducted steps 1 and 2 from section III twice; that is, we first extracted the candidates using the seeds, added the journal names to the dictionary, and extracted the candidates again.
When extracting the candidates again, we only extracted the unseen candidates.
If candidates with the same name had different scores, their left or right bigrams were different.
When a candidate had already been in the top 2,000 but was discovered to have a new higher score, the original score was overwritten with the higher score.
For the top 2,000 candidates extracted based on seeds, we only added the journal names that we decided were correct to the dictionary by hand.
By doing this, we excluded extraction failures caused by strings which are not journal names.
Therefore we could analyze the effectiveness of the distribution hypothesis in terms of only the occurrence contexts of journal names.

\begin{table}[tp]
\begin{center}
  \caption{
Seeds.
We use 10 journal names of 2 types.
These names have different notations.
}
  \begin{tabular}{| l |}
    \hline
    \multicolumn{1}{| c |}{\textbf{Journal name}} \\
    \hline
Scientific Reports \\
サイエンティフィック・リポーツ \\
サイエンティフィック・リポーツ（Scientific Reports） \\
サイエンティフィックリポーツ \\
サイエンティフィックリポーツ（Scientific Reports） \\
PLOS ONE \\
プロス・ワン \\
プロス・ワン（PLOS ONE） \\
プロスワン \\
プロスワン（PLOS ONE） \\
    \hline
  \end{tabular}
  \label{tab:Seeds}
\end{center}
\end{table}

\subsection{Dataset/journal names description}

In this section, we describe the dataset and journal names.
In our experiment, we used a collection of scientific news articles that were likely to contain journal names as the dataset; we collected scientific news articles from the past 10 years from various Japanese news websites.
Then, we used a total of 30,076 articles that were narrowed down based on the following search condition: ``学誌 OR 論文誌 OR 学術誌'' ({\it gakushi} OR {\it ronbunshi} OR {\it gakuzyutushi}, which all mean journal).
Each article contains a URL, title, body, posting date, news ID for each site, and field ID in a tab-delimited manner.
We only used the article body in our experiment.

The dataset contains many Japanese and English journal names.
The notation of the journal names is roughly divided into four patterns, as shown in TABLE~\ref{tab:Notation}.

\begin{table}[tp]
\begin{center}
  \caption{
Notation of the journal names. 
We use two names, ``Neuron'' and ``Cell Research,'' as examples.
``電子版'' means electronic edition.
}
  \begin{tabular}{| l |}
    \hline
\textbf{English name} \\
Neuron, Cell Research  \\
    \hline
\textbf{Japanese name} \\
ニューロン, セル・リサーチ \\
    \hline
\textbf{Both English and Japanese names} \\
ニューロン（Neuron）, セル・リサーチ（Cell Research） \\
    \hline
\textbf{Name with supplementary information} \\
ニューロン電子版, セル・リサーチ（電子版） \\
    \hline
  \end{tabular}
  \label{tab:Notation}
\end{center}
\end{table}

\subsection{Correct answer data and performance measures}

We used the following precision, recall, and F-measure calculations as the performance measures:
\begin{align*}
{\rm Precision} &= \frac{\rm |Correct,\ manually\ judged\ candidates|}{\rm |Candidates\ extracted\ so\ far|}, \\
{\rm Recall} &= \frac{\rm |Matching\ candidates|}{\rm |Journal\ names\ in\ the\ correct\ answer\ data|}, \\
{\rm F\ measure} &= \frac{\rm 2 \cdot Recall \cdot Precision}{\rm Recall + Precision}.
\end{align*}
To calculate the recall, we need the set of all the strings that a human could recognize as journal names as the correct answer data.
However, this data is hard to obtain.
Therefore, we obtained a list of journal names recorded on Web of Science and ScienceDirect, and those indexed by the National Diet Library.
We found the obtained journal names from the dataset by using the longest-match retrieval method and deleted other names from the list.
During retrieval, we did not distinguish between upper and lowercase letters.
Some Japanese journal names are difficult to distinguish from common terms.
For this reason, we conducted a matching search of general nouns contained in MeCab's IPA dictionary\footnote{We used mecab-ipadic-2.7.0-20070801.} and deleted any matching names of the journal names we found.
Nevertheless, since many of the names that were difficult to distinguish from general terms were contained in the list, we removed all strings containing nine characters or less.
Following the procedures mentioned above, we generated the correct answer data, which consisted of 1,037 journal names.
Although the correct answer data contains only some of the journal names in the dataset, it makes reproducible and quantitative performance evaluation of the journal name extractions possible.

\subsection{Experiment results}

The extraction performance is shown in TABLE~\ref{tab:Proposed method}.
In the first iteration, the precision value was 0.8 or higher, and the recall value was over 0.5.
From this, we found that even if a few the journal names were input as seeds, we could extract many journal names by using the left and right context features as a pattern.
However, in the second iteration, the newly extracted journal names were reduced by half, and although the precision value was reduced by 0.2, the recall value hardly improved.
These results showed that using the left and right bigrams as clues to extract the journal names were effective.
However, these results also suggested that other features were necessary to extract the journal names while maintaining a high F-measure when using the bootstrap approach.

\begin{table}[tp]
\begin{center}
  \caption{
Cumulative precision, recall, and F-measure.
}
  \begin{tabular}{l r r }
    \hline
    \hline
    \multirow{2}{*}{Metric} & \multicolumn{2}{c}{\# of iterations} \\ \cline{2-3} 
    & \multicolumn{1}{c}{1} & \multicolumn{1}{c}{2} \\
    \hline
Precision & 0.856 & 0.636 \\
Recall & 0.517 & 0.587 \\
F-measure & 0.645 & 0.611 \\
    \hline
\# of extracted journal names & 1,712 & 2,544 \\
    \hline
  \end{tabular}
  \label{tab:Proposed method}
\end{center}
\end{table}

\section{Discussion}

For the first iteration, the top three frequencies of the left and right bigrams are shown in TABLE~\ref{tab:bigrams}.
The occurrence frequencies of this table were corrected by the Simple Good-Turing method.
Even if we only check the top three bigrams, their occurrence rates occupy 50\% or more and 25\% or more of the total frequency in each of the left and right bigrams.
This numerically indicates that journal names tend to occur in a limited context.
Therefore, we can extract many journal names by accurately capturing the context features surrounding the journal names.

\begin{table}[tp]
\begin{center}
  \caption{Top three left and riight bigrams.}
  \subtable[Left bigrams]{
    \begin{tabular}{c r r }
      \hline
      \hline
      \multicolumn{1}{c}{Left bigram} & \multicolumn{1}{c}{\# of occurrences} & \multicolumn{1}{c}{Rate of total} \\
      \hline
学誌 & 14,237.146 & 0.334 \\
誌「 & 9,735.146 & 0.229 \\
ー（ & 1,112.148 & 0.026 \\
      \hline
    \end{tabular}
  }
  \subtable[Right bigrams]{
    \begin{tabular}{c r r }
      \hline
      \hline
      \multicolumn{1}{c}{Right bigram} & \multicolumn{1}{c}{\# of occurrences} & \multicolumn{1}{c}{Rate of total} \\
      \hline
」に & 4,450.283 & 0.105 \\
に発 & 4,206.283 & 0.099 \\
電子 & 2,239.284 & 0.053 \\
      \hline
    \end{tabular}
  }
  \label{tab:bigrams}
\end{center}
\end{table}

We conducted an error analysis of the experiment.
An example of an erroneous extraction in the second iteration is shown in TABLE~\ref{tab:misextraction}.
In error 1, since the context was taken from the ambiguous journal name ``ワクチン'' ({\it wakuchin}, meaning vaccine), it extracted a character string irrelevant to the journal name.
Extracting character strings unrelated to the target semantic category is generally known as a semantic drift.
In error 2, the left and right context seems to be a journal name, but unnecessary string ``で13日'' ({\it de 13 nichi}, meaning ``on the 13th'') is attached to the extracted character string.
Most of the errors that occurred were similar to those in error 2.
In order to avoid this kind of error, we should take into account not only contexts of candidate strings, but also the strings themselves.
Finally, in error 3, only a part of the journal name ``ジャーナル・オブ'' (journal of) was extracted.
A journal name is often a nested structure including another journal name as a partial character string.
For example, the journal name ``Nature Genetics'' contains another journal called ``Nature''.
When learning the occurrence context of nested journal names (e.g., ``Nature'' contained in ``Nature Genetics''), it sometimes causes the distribution hypothesis to be ineffective and extraction of incomplete parts of journal names, as seen in error 3.

\begin{table}[tp]
\begin{center}
  \caption{Some examples of error patterns.}
  \begin{tabular}{c | c  c  c }
    \hline
    \hline
    \multicolumn{1}{c |}{\#} & \multicolumn{1}{c }{Left bigram} & \multicolumn{1}{c }{Extracted string} & \multicolumn{1}{c }{Right bigram} \\
    \hline
1 & 咳） & ワクチンを & 接種 \\
2 & 学誌 & ネイチャーで13日 & に発 \\
3 & 誌「 & ジャーナル・オブ & ・パ \\
    \hline
  \end{tabular}
  \label{tab:misextraction}
\end{center}
\end{table}

\section{Conclusion}

We addressed the task of extracting journal names from Japanese scientific news articles by using a few seeds and only context features.
We constructed a character-based method and extracted journal names from Japanese scientific news articles.
This method identifies the journal names by using names' occurrence contexts in the articles.
As a result, we found that the distribution hypothesis was effective in extracting journal names from Japanese scientific news articles.
Even when using a few journal names as seeds, the results suggested that many journal names could still be extracted by using context features.
However, we sometimes extracted inadequate character strings as journal names and character strings with unnecessary strings added to them.
In addition, we often confirmed the occurrence of semantic drift.
We plan to solve these problems and create a powerful method for journal name extractions in our future work.

\section*{Appendix}

TABLE~\ref{tab:Journal list} shows some journal names extracted in this research.

\begin{table*}[tp]
\begin{center}
  \caption{
Some examples of extracted journal names.
``電子版'' and ``オンライン版'' mean electronic edition and online edition, respectively.
}
  \begin{tabular}{| l | l |}
    \hline
    \multicolumn{2}{|l|}{\textbf{English name}} \\
    \hline
Global Biogeochemical Cycles & Art Therapy \\
Infection and Immunity & Meteoritics \& Planetary Science \\
Journal of Data Mining and Knowledge Discovery & Journal of Educational Psychology \\
Molecular and Cellular Biology & Earth Systems Data Discussions \\
Aging Cell & Neurobiology of Aging \\
Environmental Science ＆ Technology Letters & Advanced Materials Interfaces \\
Journal of Crystal Growth & Science Advances \\
BMJ Open & Time and Mind \\
    \hline
    \multicolumn{2}{|l|}{\textbf{Japanese name}} \\
    \hline
ブリティッシュ・ジャーナル・オブ・ファーマコロジー & ネイチャー・レビューズ・ディジーズ・プライマーズ \\
科学通報 & ザ・サイエンティスト \\
ヒリヨン & スミソニアン・マガジン \\
バイオジオサイエンシス & ブレーンアンドランゲージ \\
サイエンスアドバンシス & コンサベーション・レターズ \\
ブレーン & ケミストリー・オブ・マテリアルズ \\ 
ヒューマン・ブレイン・マッピング & セル・ステム・セル \\
ネーチャーマテリアル & フィジカルレビューA \\
    \hline
    \multicolumn{2}{|l|}{\textbf{Both English and Japanese names}} \\
    \hline
バイオメドセントラル（BioMedCentral） & ネイチャー・プランツ（Nature Plants） \\
ブレイン・アンド・ビヘイビア（Brain and Behavior） & 王立医学会（Journal of the Royal Society of Medicine） \\
ネイチャー・ナノテクノロジー（Nature Nanotechnology） & アルコールとアルコール依存症（Alcohol and Alcoholism） \\
キャンサー（Cancer） & J・Proteome Research（プロテオームリサーチ） \\
ニュー・ジャーナル・オブ・フィジックス（New Journal of Physics） & ヒューマン・リプロダクション（Human Reproduction） \\
Urology（泌尿器学） & ナチュラル・ハザーズ（Natural Hazards） \\
Journal of Public Health（公衆衛生ジャーナル） & セル・プレス（Cell Press） \\
Coral Reefs（コーラルリーフス） & Cell Metabolism（細胞代謝） \\
    \hline
    \multicolumn{2}{|l|}{\textbf{Name with supplementary information}} \\
    \hline
サイエンティフィック・リポーツのオンライン版 & ネイチャーニューロサイエンス電子版 \\
ジャーナル・オブ・ヒューマン・ジェネティクス電子版 & ヨーロピアン・ジャーナル・オブ・フィジックス電子版 \\
モレキュラー・システムズ・バイオロジー電子版 & アース・アンド・プラネタリー・サイエンス・レターズ電子版 \\
サーキュレーション・リサーチ（電子版） & ニューイングランド・ジャーナル・オブ・メディシン電子版 \\
プラント・セルの電子版 & アストロフィジカル・ジャーナル・レターズ電子版 \\
デベロップメンタルセル電子版 & アプライド・フィジックス・エクスプレス電子版 \\
ネイチャー・バイオテクノロジーの電子版 & ネイチャージオサイエンス電子版 \\
ネイチャー・イムノロジー電子版 & ネイチャー・イミュノロジー電子版 \\
    \hline
  \end{tabular}
  \label{tab:Journal list}
\end{center}
\end{table*}

\balance{}
\bibliographystyle{IEEEtran}
\bibliography{IEEEabrv,fullpaper}

\end{document}